\documentclass{article}
\usepackage[final,nonatbib]{templ/nips_2017}
\usepackage[numbers]{natbib}
\usepackage{amsmath,amssymb,amsfonts,amsthm}
\usepackage[pdftex]{graphicx}
\usepackage[utf8]{inputenc} 
\usepackage[T1]{fontenc}    
\usepackage[draft=false, pagebackref=false,breaklinks=true,colorlinks,bookmarks=false,pdftex]{hyperref} 
\hypersetup{hidelinks,colorlinks=false,linkbordercolor={1 1 1}} 
\usepackage{url}            
\usepackage{booktabs}       
\usepackage{amsfonts}       
\usepackage{nicefrac}       
\usepackage{microtype}      
\usepackage{cvpr-abbriv}    
\usepackage{xcolor} 
\usepackage{stmaryrd}       
\usepackage[capitalise,nameinlink]{cleveref} 
\theoremstyle{definition}
\newtheorem{Definition}{Definition}
\theoremstyle{remark}
\newtheorem{Remark}{Remark}
\newtheorem{Observation}{Observation}
\newtheorem{Example}{Example}
\theoremstyle{plain}

\newcommand{\koerper}[1]{\mathbb{#1}}

\newcommand{\E}{{\koerper{E}}}

\providecommand{\mvec}[1]{\boldsymbol{#1}}
\newcommand{\x}{{\mvec{x}}}
\newcommand{\y}{{\mvec{y}}}

\providecommand{\abs}[1]{\lvert#1\rvert}

\providecommand{\scalp}[2]{\left<#1,#2\right>}
\newcommand{\<}{\langle}
\renewcommand{\>}{\rangle}

\graphicspath{{images/}}
\DeclareGraphicsExtensions{.pdf,.jpeg,.png}
\title{Generative learning for deep networks}
\author{
	Boris Flach \\
	\texttt{flachbor@cmp.felk.cvut.cz}\\
  \And
  Alexander Shekhovtsov \\
	\texttt{shekhovtsov@gmail.com}\\
  \And
  Ondrej Fikar \\
	\texttt{fikarond@fel.cvut.cz}\\
	\AND
	\\[-20pt]
	Czech Technical University in Prague
}

\begin{document}

\maketitle

\begin{abstract}
Learning, taking into account full distribution of the data, referred to as generative, is not feasible with deep neural networks (DNNs) because they model only the conditional distribution of the outputs given the inputs. Current solutions are either based on joint probability models facing difficult estimation problems or learn two separate networks, mapping inputs to outputs (recognition) and vice-versa (generation). We propose an intermediate approach. First, we show that forward computation in DNNs with logistic sigmoid activations corresponds to a simplified approximate Bayesian inference in a directed probabilistic multi-layer model. This connection allows to interpret DNN as a probabilistic model of the output and all hidden units given the input. Second, we propose that in order for the recognition and generation networks to be more consistent with the joint model of the data, weights of the recognition and generator network should be related by transposition. We demonstrate in a tentative experiment that such a coupled pair can be learned generatively, modelling the full distribution of the data, and has enough capacity to perform well in both recognition and generation. 
\end{abstract}
\section{Introduction}
Neural networks with logistic sigmoid activations may look very similar to Bayesian networks of the same structure with logistic conditional distributions, aka {\em sigmoid belief networks}~\cite{Neal:1992}. However, hidden units in NNs are deterministic and take on real values while hidden units in Bayesian networks are binary random variables with an associated distribution. Given enough capacity and training data, both models can estimate the posterior distribution of their output arbitrary well. Besides somewhat different modelling properties, there is a principled difference that with stochastic models it is possible to pose, at least theoretically, a number of inference and learning problems with missing data by marginalizing over latent and unobserved variables. Unfortunately, even forward inference in Bayesian networks requires methods such as sampling or optimization of a variational approximation. Likewise, in more tightly coupled graphical models such as {\em deep Boltzmann machine} (DBM)~\cite{salakhutdinov09a} or deep belief networks (DBN)~\cite{Hinton:NC2006,Lee:ICML2009} practically all computations needed e.g.~for computing marginal and posterior probabilities are not tractable in the sense that approximations typically involve sampling or optimization.

In this paper we propose stacked (deep) conditional independent model (DCIM). There are two views how the model can be defined. One, as a Bayesian network with logistic conditional distributions. The other, just assuming that conditional probabilities of a general Bayesian network factor over the parent nodes up to a normalising factor. With binary units this necessary implies that conditional probabilities are logistic. It is noteworthy that we find the same form of conditional probabilities in most of the neural probabilistic models: restricted Boltzmann machine, DBM~\cite{salakhutdinov09a}, DBN~\cite{Hinton:NC2006}, etc. When the units are arranged in layers, as in a typical DNN, the layered Bayesian network can be viewed as a Markov chain with a state space of all variables in a layer. In this view, all necessary  assumptions can be summarized in one property, termed {\em strong conditional independence} that the forward conditional transition probabilities between the layers factorize over the dimensions of both the input and output state spaces of the transition up to a normalising factor. Making only this assumption, we show that a simple approximate Bayesian inference in DCIM recovers main constructive elements of DNNs: sigmoid and softmax activations with its real-valued variables corresponding to expectations of binary random variables. With this interpretation we can view DNN as a way of performing approximate inference very efficiently. To our knowledge, this relationship has not been established before. Under this approximation conditional likelihood learning of a DCIM is equivalent to that of DNN and can be performed by back-propagation.

Our second objective is to propose an alternative, generative learning approach for DNNs. In a number of recent works~\cite{KingmaW13,Rezende:ICML2014,Salakhutdinov:ARSA2015,Mnih-2014} and in the prior work~\cite{HintonDayanFreyEtAl95,Hinton:NC2006} a pair of deterministic recognition (encoder) and a stochastic generator (decoder) networks is trained. Let us denote $\x^0$ the input of the recognition network, to be specific, an image, and $\x^d$ (random variables at layer $d$) the output of the recognition network, the latent state. Although the two networks often are taken to have a symmetric structure~\cite{HintonDayanFreyEtAl95,KingmaW13}, their parameters are decoupled. The stochastic generator network can typically only generate samples, but cannot directly evaluate its posterior distribution or the gradient thereof, requiring variational or sampling-based approximations. The methods proposed in~\cite{KingmaW13,Rezende:ICML2014} assume that samples generated from the same latent state $x^d$ must fall close together in the image space. This prohibits using categorical latent spaces such as digit class in MNIST because digits of the same class naturally may look differently. Instead, a model's own continuous latent space is used such that fixing a point in it defines both the class and the shape of the digit. Works~\cite{KingmaW13,Rezende:ICML2014} are thus restricted to unsupervised learning.

Given full training data pairs of $(\x^0, \x^d)$, the recognition network could learn a distribution $p(\x^d \mid \x^0)$ and the generator network could in principle learn a distribution $q(\x^0 \mid \x^d)$. With our link between DNN and DCIM, we ask the question when the two DCIMs, modelling the two conditional distributions $p(\x^0 \mid \x^d)$ and $q(\x^d \mid \x^0)$, are consistent, i.e., correspond to some implicitly modelled joint distribution $p^*(\x^{0,\dots, d})$ of the data and all hidden layers. It is the case when $p(\x^{1,\dots d} \mid \x^0)/ q(\x^{0,\dots d-1}\mid \x^d) = A(\x^0)B(\x^d)$ for some functions $A,B$. While this cannot be strictly satisfied with our strong conditional independence assumptions, we observe that most of the terms in the ratio cancel when we set the weights of recognition network $p$ to be transposed of those of network $q$. Both models therefore can be efficiently represented by one and the same DNN, share its parameters and can be learned simultaneously by using an estimator similar to pseudo-likelihood. We further use the link between DNN and DCIM to approximately compute the posterior distribution in the generator network $q(\x^d \mid \x^k)$ given a sample of an inner layer $\x^k$. The approximation is reasonable when this posterior is expected to have a single mode, such as when reconstructing an image from lower level features. We thus can fully or partially avoid sampling in the generator model. We demonstrate in a tentative experiment that such a coupled pair can be learned simultaneously, modelling the full distribution of the data, and has enough capacity to perform well in both recognition and generation.

\paragraph{Outline}
In~\cref{sec:approx} we consider strongly conditional independent model with two layers and derive the NN from it.
This model will serve as a building block of DCIM formally introduced in~\cref{sec:DCIM}.
In~\cref{subsec:gen_learn} we consider coupled pairs of DCIMs and propose a novel approach for generative learning.
In~\cref{sec:related} we discuss more connections with related work. In~\cref{sec:experiments} we propose our proof-of-concept experiments of generative learning using only DNN inference or sampling only a single layer.
%
\section{Deep conditional independent models}
\subsection{An approximation for conditional independent models}\label{sec:approx}
Let $\mvec{x}=(x_1,\ldots,x_n)$ and $\mvec{y}=(y_1,\ldots,y_m)$ be two collections of discrete random variables with a joint distribution $p(\mvec{x}, \mvec{y}) = p(\mvec{y} \mid \mvec{x}) \: p(\mvec{x})$. 
The conditional distribution $p(\mvec{y} \mid \mvec{x})$ is {\em strongly conditional independent} if it factors as $p(\mvec{y} \mid \mvec{x})  = \frac{1}{Z(\mvec{x})} \prod_{i,j}g_{ij}(x_i, y_j)$. 
Without loss of generality, it can be written as
%
\begin{equation} \label{eq:cim}
 p(\mvec{y} \mid \mvec{x}) = \frac{1}{Z(\mvec{x})} 
 \exp \sum_{i,j} u_{ij}(x_i, y_j) = 
 \prod_{j=1}^{m} p(y_j \mid \mvec{x}) = 
 \prod_{j=1}^{m} \frac{1}{Z_j(\mvec{x})} 
 \exp \sum_{i=1}^{m} u_{ij}(x_i, y_j) .
\end{equation}
The functions $u_{ij}$ are arbitrary and $Z_j(\x)$ denote the corresponding normalization constants.
While sampling from $p(\mvec{y} \mid \mvec{x})$ is easy, computing $p(\mvec{y})$ by marginalizing over $\mvec{x}$ is not tractable even if $p(\mvec{x})$ factorizes over the components of $\mvec{x}$.

We consider the following approximation for the marginal distribution $p(\y)$:
\begin{equation} \label{eq:pyapprox}
 p(\mvec{y}) \approx \frac{1}{Z} \exp 
 \sum_{ij} \sum_{x_i} p(x_i) u_{ij}(x_i, y_j) .
\end{equation}
One of the possible ways to derive it, is to consider Taylor expansions for moments of functions of random variables. By using a linear embedding $\mvec{\Phi}$ such that $\sum_{i,j}u_{ij}(x_i,y_j) = \scalp{\mvec{\Phi}(\mvec{x},\mvec{y})}{\mvec{u}}$, where $\mvec{u}$ is a vector with components $u_{i,j}(k,l)$, we can write $p(\mvec{y} \mid \mvec{x})$ in the form
\begin{equation}
 p(\mvec{y} \mid \mvec{x}) = \frac{\exp \scalp{\mvec{\Phi}(\mvec{x},\mvec{y})}{\mvec{u}}}%
 {\sum_{\mvec{y}'} \exp \scalp{\mvec{\Phi}(\mvec{x},\mvec{y}')}{\mvec{u}}} .
\end{equation}
The marginal distribution $p(\y)$ is obtained from
\begin{equation}
 \E_{\mvec{x}\sim p(\mvec{x})} p(\mvec{y} \mid \mvec{x}) = 
 \E_{\mvec{x}\sim p(\mvec{x})} 
 \frac{\exp \scalp{\mvec{\Phi}(\mvec{x},\mvec{y})}{\mvec{u}}}%
 {\sum_{\mvec{y}'} \exp \scalp{\mvec{\Phi}(\mvec{x},\mvec{y}')}{\mvec{u}}} .
\end{equation}
Taking the first order Taylor expansion w.r.t.~the random variable $\mvec{\Phi}(\mvec{x},\mvec{y})$ for fixed $\y$ around 
\begin{equation}
 \overline{\Phi}(\mvec{y}) = \E_{\mvec{x}\sim p(\mvec{x})} \mvec{\Phi}(\mvec{x},\mvec{y})
\end{equation}
and computing its expectation w.r.t.~$\mvec{x}$ gives \eqref{eq:pyapprox} as the constant term, while the first order term vanishes.

\begin{Example} \label{ex:sigmoid}
 Assume that random variables in the two collections $\x=(x_1,\ldots,x_n)$ and $\y=(y_1,\ldots,y_m)$ are $\{0,1\}$-valued. Any function $u_{ij}(x_i, y_j)$ of binary variables can be written as $u_{ij}(x_i, y_j) =  y_j W_{ji}x_i + b_j y_j + c_i x_i + d$. Terms $c_i x_i + d$ cancel in the normalization of $p(\mvec{y} \mid \mvec{x})$ and thus can be omitted. The approximation \eqref{eq:pyapprox} reads then $\hat{p}(\mvec{y}) = \prod_{j} \hat{p}(y_j)$ with
 \begin{equation}\label{approx-binary1}
  \hat{p}(y_j) = \frac{ e^{(\scalp{\mvec{w}^j}{\bar{\mvec{x}}} + b_j)y_j}}{
	e^{\scalp{\mvec{w}^j}{\bar{\mvec{x}}} + b_j} + 1},
  \end{equation}
 where $\mvec{w}^j$ denotes the $j$-th row of the matrix $W$ and $\bar{\mvec{x}} = \E_{\mvec{x}\sim p(\mvec{x})} \x$. This in turn leads to the following approximation for the expectation of $\mvec{y}$:
 \begin{equation}
  \bar{y}_j = \hat p(y_j{=}1) 
	=  \frac{1}{1 + e^{-\<\mvec{w}^j,\x\> - b_j}
	} = \mathcal{S}(\<\mvec{w}^j,\bar \x\> + b_j),
 \end{equation}
 where $\mathcal{S}$ is the logistic (sigmoid) function.

 If $\pm 1$ encoding is used instead to represent states of $\x$ and $\y$, then the corresponding expressions read
  \begin{equation}\label{approx-binary2}
  \hat p(y_j) = \frac{\exp\bigl[ 
  y_j \bigl( \scalp{\mvec{w}^j}{\bar{\mvec{x}}} + b_j \bigr)\bigr]}
  {2 \cosh \bigl[\scalp{\mvec{w}^j}{\bar{\mvec{x}}} + b_j\bigr]} ,
 \end{equation}
 and
  \begin{equation}
  \bar{y}_j = \tanh [ \scalp{\mvec{w}^j}{\bar{\mvec{x}}} + b_j ] .
 \end{equation}
\end{Example}

\begin{Remark}
Let $p(\mvec{x}, \mvec{y})$ be a joint model such that $p(\mvec{y} \mid \mvec{x})$ is strongly conditional independent as above. Then $p(\mvec{x} \mid \mvec{y})$ is not in general conditional independent. 
If both conditional distributions are strongly conditional independent, then the joint model is a restricted Boltzmann machine or, what is the same, an MRF on a bipartite graph.
\end{Remark}
\subsection{Deep conditional independent models}\label{sec:DCIM}
Let $\mathcal{X} = \bigl(\mvec{x}^0,\mvec{x}^1,\ldots,\mvec{x}^d\bigr)$ be a sequence of collections of binary valued random variables 
\begin{equation} \label{eq:seq_coll}
 \mvec{x}^k = \bigl\{ x_i^k = \pm 1 \bigm | i = 1,\ldots, n_k \bigr\} .
\end{equation}
The next observation highlights the difficulties that we run into if considering the deep Boltzmann machine model.
\begin{Observation} \label{obs:non-cim}
 Let us assume that the joint distribution $p(\mathcal{X})$ is a deep Boltzmann machine, i.e.~an MRF on a layered, $d$-partite graph. Then neither the forward conditionals $p(\mvec{x}^k \mid \mvec{x}^{k-1})$ nor the backward conditionals $p(\mvec{x}^{k-1} \mid \mvec{x}^{k})$ are (conditionally) independent. This can be seen as follows. The joint distribution of such a model can be written w.l.o.g.~as
 \begin{equation}
  p(\mathcal{X}) = \frac{1}{Z} \exp \Bigr[
  \sum_{k=0}^{d} \scalp{\mvec{b}^k}{\mvec{x}^k} + 
  \sum_{k=0}^{d} \scalp{\mvec{x}^k}{W^k \mvec{x}^{k-1}} \Bigr]
 \end{equation}
 and it follows that
 \begin{equation}
   p(\mvec{x}^k \mid \mvec{x}^{k-1}) = 
  \frac{F(\mvec{x}^k)}{Z(\mvec{x}^{k-1})} \exp \Bigr[
  \scalp{\mvec{b}^k}{\mvec{x}^k} + 
  \scalp{\mvec{x}^k}{W^k \mvec{x}^{k-1}} \Bigr] 
 \end{equation}
 where $F(\mvec{x}^k)$ results from marginalisation over $\mvec{x }^{k+1},\ldots,\mvec{x}^d$ and can be arbitrarily complex.
\end{Observation}
We will therefore consider a similar, but different class, for which the forward conditionals are (conditionally) independent.
\begin{Definition} \label{def:dcim}
 A joint distribution $p(\mathcal{X})$ for a sequence of collections of binary valued random variables \eqref{eq:seq_coll} is a deep conditional independent model (DCIM) if it has the form
 \begin{equation} \label{eq:DCIM}
  p(\mvec{x}^0,\ldots,\mvec{x}^n) = p(\mvec{x}^0) 
  \prod_{k=1}^{d} p(\mvec{x}^k \mid \mvec{x}^{k-1})
 \end{equation}
 and its forward conditionals $p(\mvec{x}^k \mid \mvec{x}^{k-1})$ are strongly conditional independent, i.e.~have the form \eqref{eq:cim} for all $k=1,2,\ldots,d$.
\end{Definition}
A DCIM can be thus seen as an inhomogeneous Markov chain model with high dimensional state spaces $\mvec{x}^k \in \mathcal{X}^k$, however with an yet unspecified distribution for $\mvec{x}^0$. Given such a model and a realisation of $\mvec{x}^0$, the posterior conditional distribution $p(\mvec{x}^d \mid \mvec{x}^0)$ can be computed by applying the approximation \eqref{eq:pyapprox} recursively (see Example.~\ref{ex:sigmoid}). This leads to the recursion
\begin{equation}
 \bar{x}_i^k = \tanh \bigl(
 \scalp{\mvec{w}^k_i}{\bar{\mvec{x}}^{k-1}} + b^k_i \bigr)
 \text{,\hspace{.5em}$k=1,2\ldots,d-1$},
\end{equation}
where $\bar{\mvec{x}}^k$ denotes the expectation of $\mvec{x}^k$ w.r.t.~$p(\mvec{x}^k \mid \mvec{x}^0)$. Finally, we obtain
\begin{equation}
 p(x^d_i \mid \mvec{x}^0) = 
 \frac{\exp(x^d_i a^d_i)}{2 \cosh(a^d_i)} 
 \text{\hspace{.5em}with\hspace{.5em}}
 a^d_i = \scalp{\mvec{w}^d_i}{\bar{\mvec{x}}^{d-1}} + b^d_i .
\end{equation}
It is obvious that the (approximate) computation of $p(\mvec{x}^d \mid \mvec{x}^0)$ for a DCIM is exactly the forward computation for a corresponding DNN. 
Hence, discriminative learning of a DCIM by maximizing the conditional likelihood $p(\x^d \mid \x^0)$ of training samples $(\x^d, \x^0)$ means to learn the corresponding DNN with the same loss.
\par
With 0-1 state variables the model is equivalent to a Bayesian network (a directed graphical model), in which the set of parents of a variable $x^k_j$ is the preceding layer $\x^{k-1}$ (or its subset, depending on the structure of the weights) and the conditional distributions are of the form~\eqref{eq:cim}, \ie, logistic. Such networks were proposed by~\citet{Neal:1992} under the name sigmoid belief networks as a simpler alternative to Boltzmann machines. When we derive the model from the strong conditional independence assumption, it establishes a link to deep Boltzmann machines~\cite{salakhutdinov09a}, which are graphical models factorizing over a $d$-partite graph.
\subsection{Generative learning of deep networks} \label{subsec:gen_learn}
To learn a DCIM generatively from a given i.i.d.~training data $\mathcal{T}$, it is necessary to specify the joint model $p(\mvec{x}^0,\ldots,\mvec{x}^d)$ and to choose an appropriate estimator. In theory, the DCIM model can be completed to a joint model by specifying a prior distribution $p(\mvec{x}^0)$ over images and then the maximum likelihood estimator can be applied. It is however not realistic to model this complex multi-modal distribution in a closed form. We propose to circumvent this problem by applying the following bidirectional conditional likelihood estimator:
\begin{equation} \label{eq:pplik}
 \frac{1}{\abs{\mathcal{T}}} 
 \sum_{(\mvec{x}^0,\mvec{x}^d)\in \mathcal{T}} \bigl[
 \log p_{\theta}(\mvec{x}^d \mid \mvec{x}^0) + 
 \log p_{\theta}(\mvec{x}^0 \mid \mvec{x}^d) \bigr] 
 \rightarrow \max_{\theta},
\end{equation}
because it avoids to model $p(\mvec{x}^0)$ and $p(\mvec{x}^d)$ explicitly. 

From the definition of a DCIM, and particularly from the fact that a DCIM is also a Markov chain model, follows that its reverse conditional distribution factorises similar as \eqref{eq:DCIM}, i.e.
\begin{equation}
 p(\mvec{x}^0,\ldots,\mvec{x}^{d-1} \mid \mvec{x}^d) = 
 \prod_{k=0}^{d-1} p(\mvec{x}^k \mid \mvec{x}^{k+1}) .
\end{equation}
This holds for any $p(\mvec{x}^0)$ completing the DCIM to a joint distribution. Unfortunately, however, the reverse conditionals $p(\mvec{x}^k \mid \mvec{x}^{k+1})$ of a DCIM are {\em not} (conditionally) independent. This follows from an argument similar to the one used in Observation~\ref{obs:non-cim}. In order to resolve the problem, we propose to consider pairs of tightly connected DCIMs: a forward model and a reverse (i.e.~backward) model
\begin{equation} \label{eq:twins}
 p(\mvec{x}^d,\ldots,\mvec{x}^1 \mid \mvec{x}^0 ) = 
 \prod_{k=1}^{d} p(\mvec{x}^k \mid \mvec{x}^{k-1}) 
 \text{\hspace{1em}and\hspace{1em}}
 q(\mvec{x}^0,\ldots,\mvec{x}^{d-1} \mid \mvec{x}^d) = 
 \prod_{k=0}^{d-1} q(\mvec{x}^k \mid \mvec{x}^{k+1})
\end{equation}
such that the forward conditionals of $p$ and the backward conditionals of $q$ are strongly (conditional) independent. Such a pair would give rise to a single joint distribution if
\begin{equation} \label{eq:wish}
 \frac{p(\mvec{x}^d,\ldots,\mvec{x}^1 \mid \mvec{x}^0 )}%
 {q(\mvec{x}^0,\ldots,\mvec{x}^{d-1} \mid \mvec{x}^d)} = 
 \frac{A(\mvec{x}^d)}{B(\mvec{x}^0)}
\end{equation}
holds for any realisation of $\mathcal{X}$ for some functions $A,B$. We already know that this is impossible to achieve while retaining conditional independence of both, the forward and backward conditionals. Therefore, we propose to choose the parameters for two models such that the equation is fulfilled as close as possible. Substituting \eqref{eq:cim} for the binary case in the lhs of \eqref{eq:wish}, gives
\begin{equation}
 \frac{\prod_{k=0}^{d-1}\tilde{Z}(\mvec{x}^{k+1})}{\prod_{k=1}^{d} Z(\mvec{x}^{k-1})}
 \frac{\exp \Bigl[
  \sum_{k=1}^{d} \scalp{\mvec{x}^k}{W^k \mvec{x}^{k-1} + \mvec{b}^k} \Bigr]}%
  {\exp \Bigl[
  \sum_{k=0}^{d-1} \scalp{\mvec{x}^k}{V^k \mvec{x}^{k+1} + \mvec{a}^k} \Bigr]}  
  \overset{?}{=}
  \frac{A(\mvec{x}^d)}{B(\mvec{x}^0)} ,
\end{equation}
where $Z_k$ and $\tilde{Z}_k$ denote the partition functions of $p(\mvec{x}^k \mid \mvec{x}^{k-1})$ and $q(\mvec{x}^k \mid \mvec{x}^{k+1})$ respectively. It is readily seen that all terms in the exponentials cancel out if $(V^{k-1})^T = W^k$ holds for the weights and $\mvec{b}^k = \mvec{a}^{k-1}$ holds for the biases. The remaining, not cancelling terms in the lhs of \eqref{eq:wish} are then the partition functions of the conditional distributions.

Summarising, a pair of such DCIMs share the same structure as well as their weights and biases. They are therefore represented by a single DNN and can be learned simultaneously by the estimator \eqref{eq:pplik}, which now reads as
\begin{equation} \label{eq:pplik2}
 \frac{1}{\abs{\mathcal{T}}} 
 \sum_{(\mvec{x}^0,\mvec{x}^d)\in \mathcal{T}} \bigl[
 \log p_{\theta}(\mvec{x}^d \mid \mvec{x}^0) + 
 \log q_{\theta}(\mvec{x}^0 \mid \mvec{x}^d) \bigr] 
 \rightarrow \max_{\theta}.
\end{equation}

Since both, the forward and backward conditionals are strongly independent, approximation \eqref{eq:pyapprox} can be applied for computing the probabilities $p(\mvec{x}^d \mid \mvec{x}^0)$ and $q(\mvec{x}^0 \mid \mvec{x}^d)$.
\begin{Remark}
 If the model consists of one layer only ($d=1$), then the pair of $p(\mvec{x}^1 \mid \mvec{x}^0)$ and $q(\mvec{x}^0 \mid \mvec{x}^1)$ define a single joint distribution, which is an RBM, i.e. an MRF on a bipartite graph. The estimator \eqref{eq:pplik} becomes the pseudo-likelihood estimator.
\end{Remark}
\subsection{Relations to other models}\label{sec:related}
We briefly discuss relations between the proposed model and other models not mentioned in the introduction. 

The connection between Bayesian networks and neural networks with injected noise, used in~\cite{Kingma:2014,Rezende:ICML2014} is different from ours. It relates two equivalent representations of stochastic models, while we relate stochastic and deterministic models.

Similar to our work, \citet{Hinton:NC2006} uses a constraint that generator and recognition networks are related by transposition of weights in the unsupervised, layer-by-layer pre-training phase, motivated by their study of deep Boltzmann machine. For the supervised fine-tuning the two models are decoupled.

The proposed model is of course related to classical auto-encoders \cite{Rumelhart:Nature1986} at least technically. Our learning approach is generative, but in contrast to auto-encoders it is supervised. Moreover, the ``decoding'' part (the reverse model) is in our case constrained to have the same parameters as the ``encoding'' part (the forward model).

Our model and the proposed generative learning approach are undoubtedly related to generative adversarial networks (GAN), \cite{Goodfellow:NIPS2014},\cite{Radford:ICLR2016} proposed by Goodfellow et al. Similar to them, the reverse part of our model aims at generating data. But in contrast to them our model uses no specific ``noise space'' and conditions the image distribution on the class. The randomness in the generating part comes solely through the reverse Markov chain. We believe that this should suffice for modelling rich distributions in the input space. It is however to expect that the approximation \eqref{eq:pyapprox} used for learning might impose limitations on the expressive power of this model part as compared with GANs (see experiments). Another difference is that our model is learned jointly with a ``twin'', i.e. the forward part, rather than in competition with an ``adversarial''.  

\section{Experiments}\label{sec:experiments}

\subsection{MNIST dense}
\begin{figure}[htb]
  {\centering
  \includegraphics[width=0.45\textwidth]{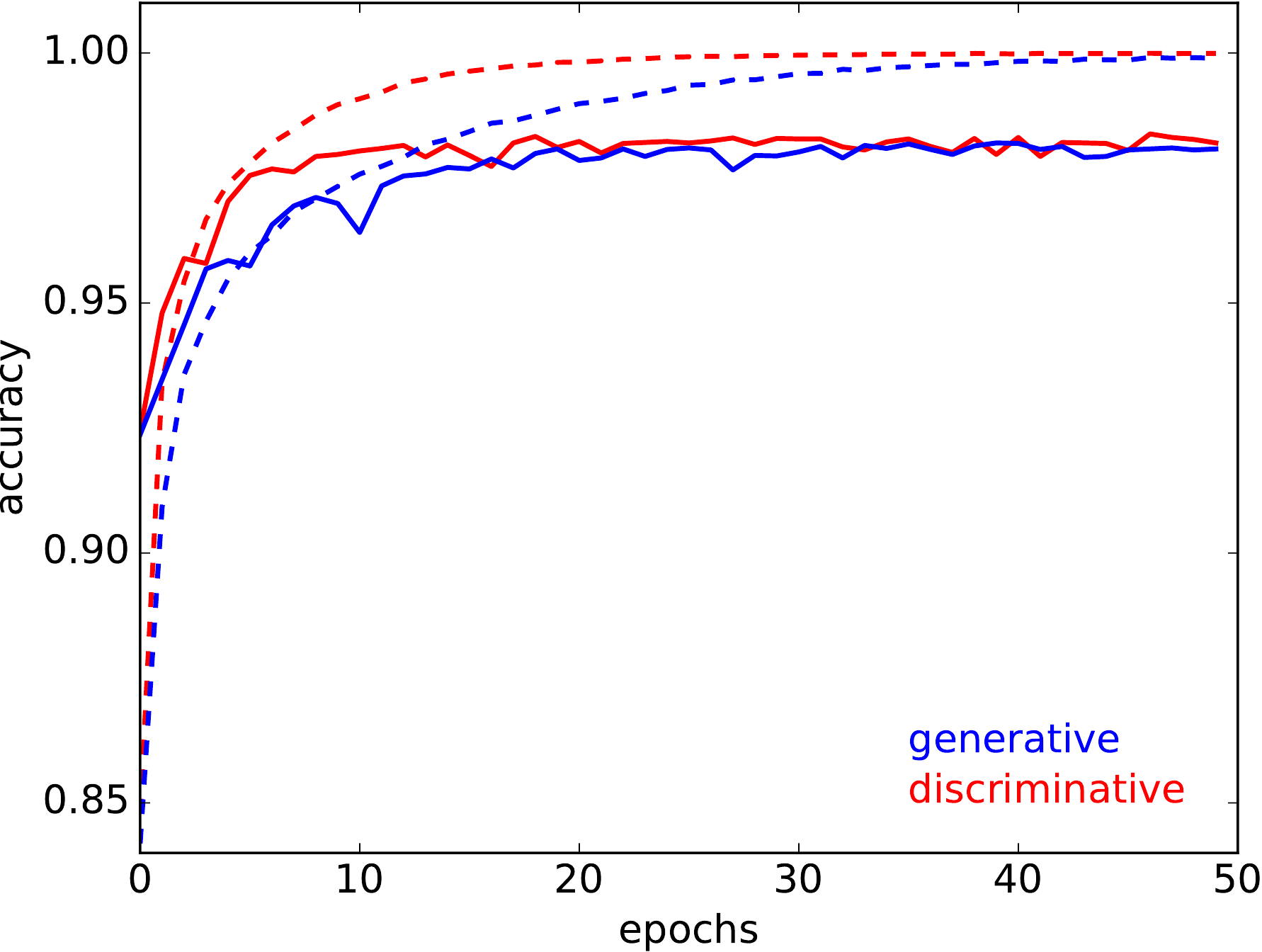} \hfil
  \includegraphics[width=0.45\textwidth]{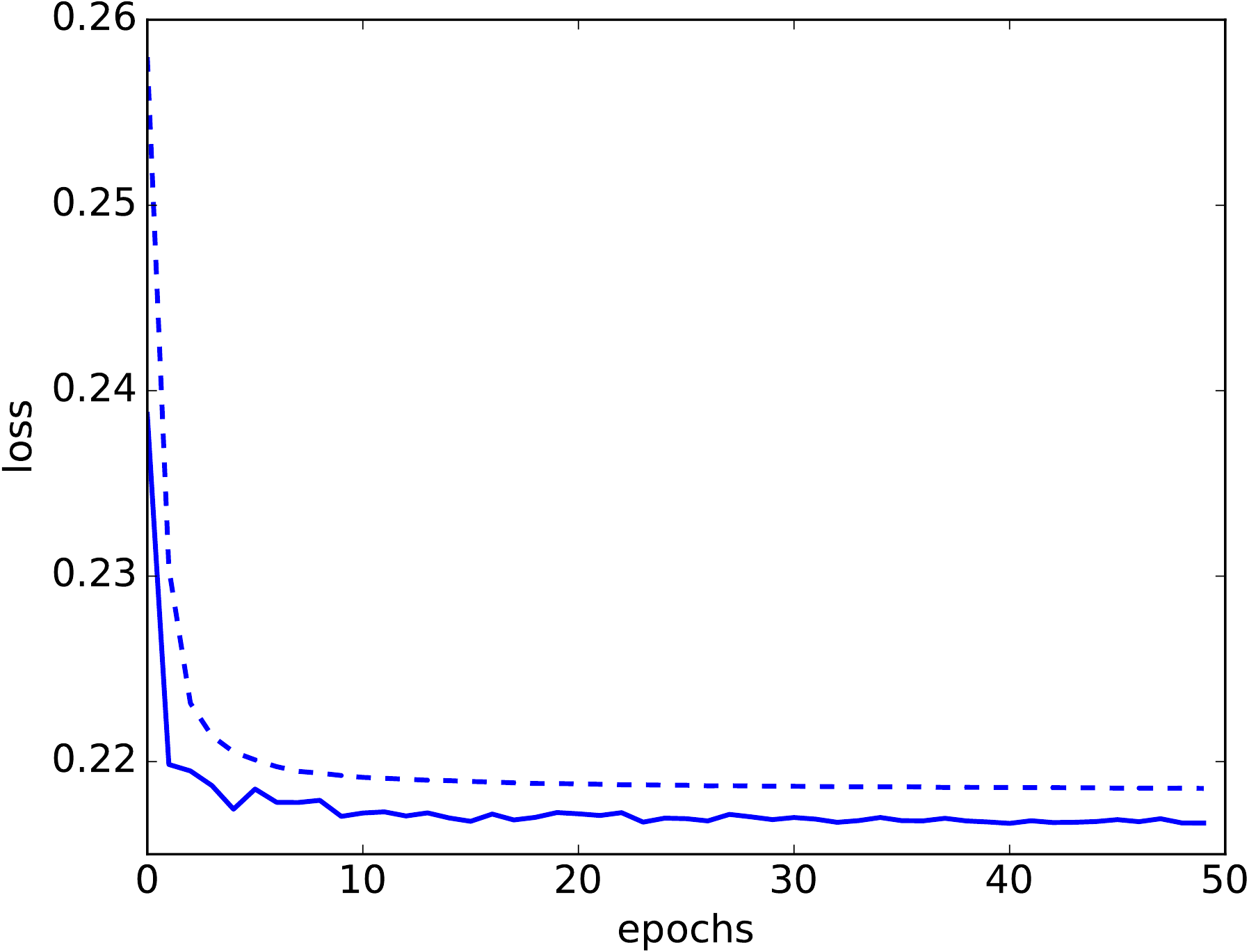}\\
  }
  \caption{Learning simple network on MNIST. 
  Left: accuracy for discriminative and generative learning (dashed - training, solid: validation).
  Right: loss of the reverse model in generative learning (dashed - training, solid: validation).}
   \label{fig:mnist-mlp-learning}
\end{figure}

The first experiment is a simple proof of concept. We trained a small network (three fully connected layers with 512, 512, 10 nodes) on the standard MNIST dataset \cite{LeCun:MNIST2010} for both learning approaches: the discriminative and the generative one. Sigmoid activation was chosen for all but the last layer, for which we used soft-max activation, i.e.~considering its nodes as categorical variable. A standard optimiser without drop-out and any other regularisers was used for searching the maximum of the objective function. Fig.~\ref{fig:mnist-mlp-learning} shows the training and validation accuracies for both approaches as functions of the iterations. It is clearly seen that the combined objective \eqref{eq:pplik} of the generative learning imposes only a rather insignificant drop of the classification accuracy. Fig.~\ref{fig:mnist-mlp-learning} also shows the training and validation loss of the the backward model.

\begin{figure}[htb]
  {\centering
  \includegraphics[width=0.4\textwidth]{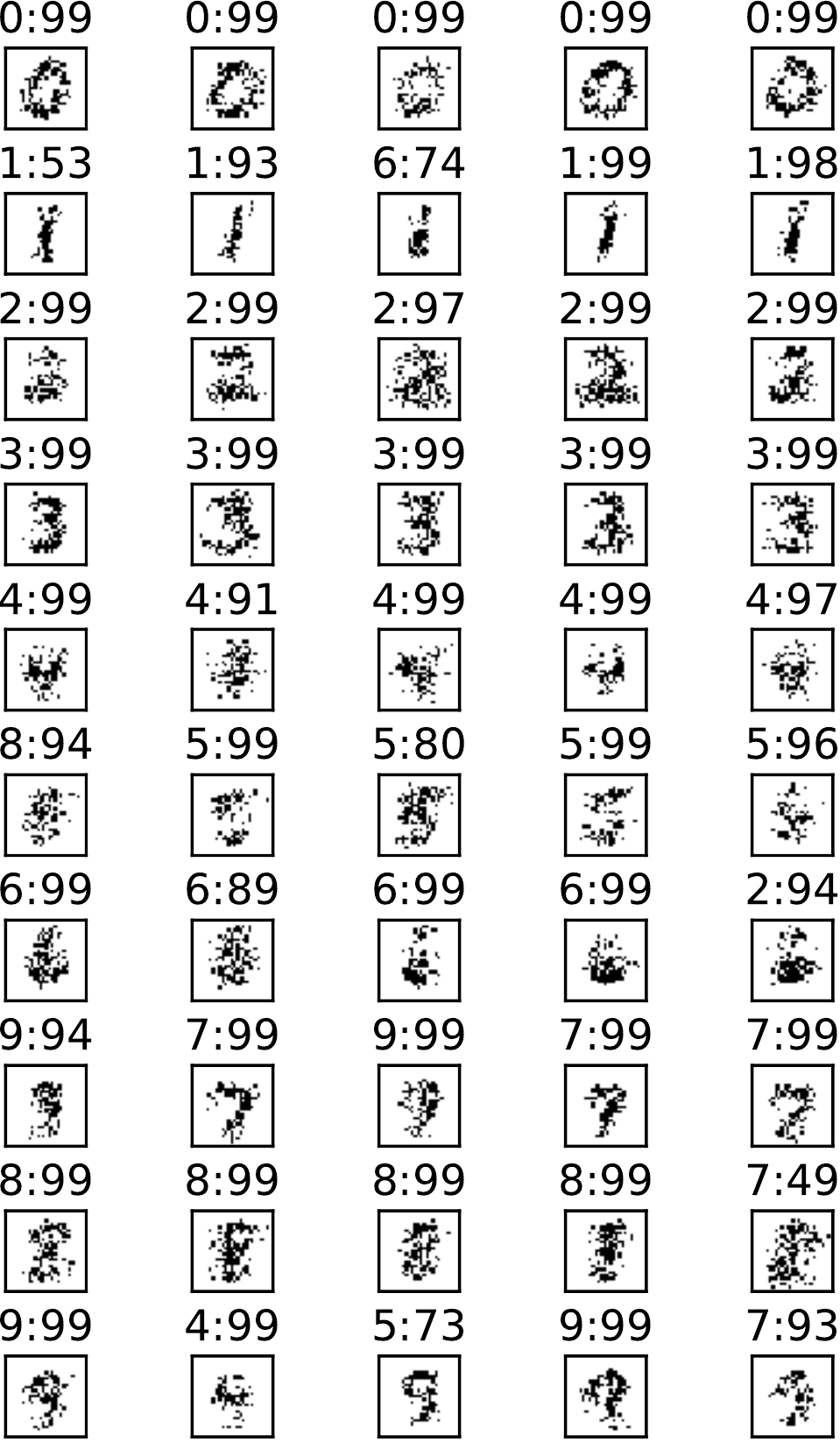} \hspace{0.06\textwidth}
  \includegraphics[width=0.4\textwidth]{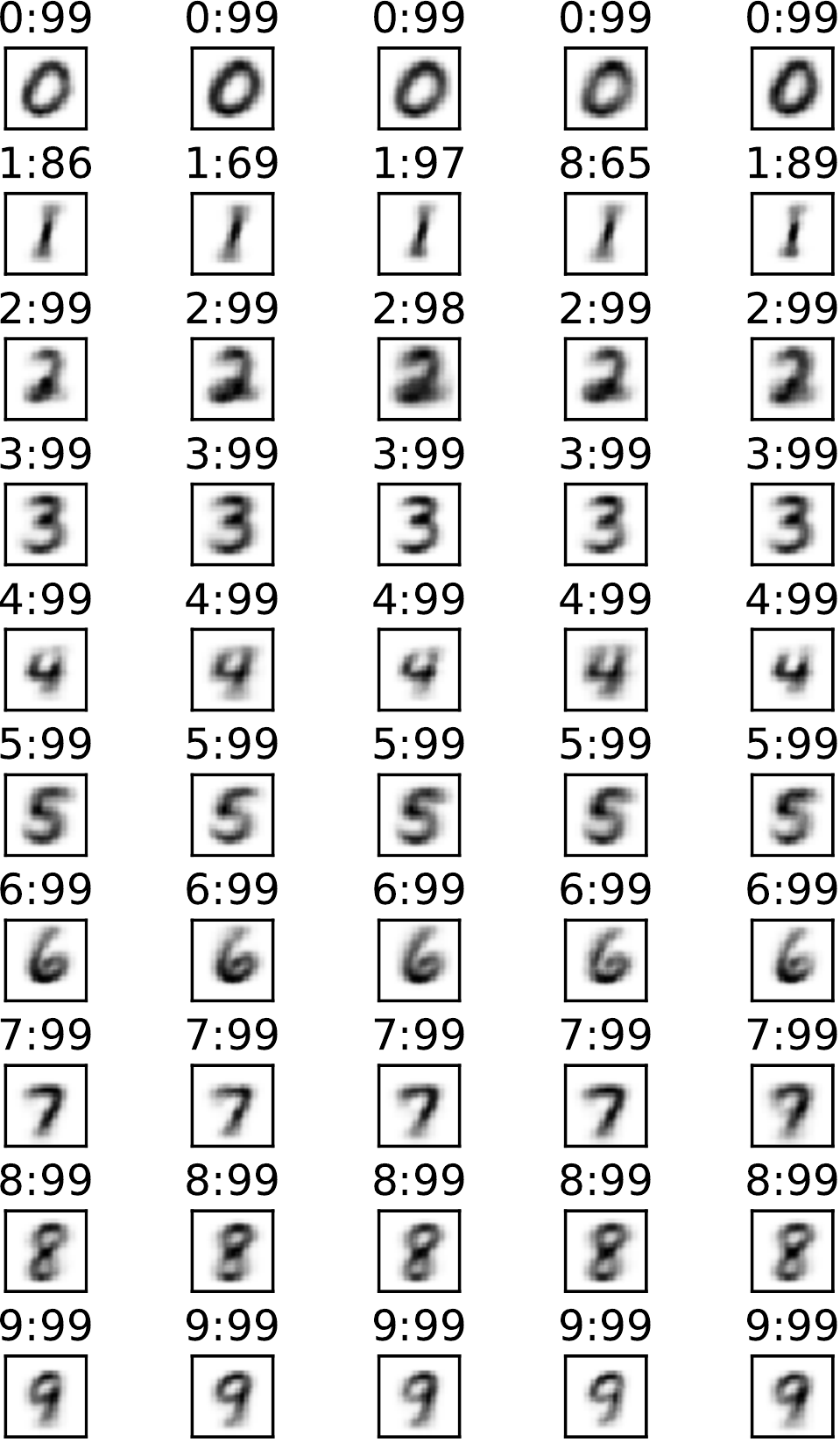}\\
  }
  \caption{Images sampled form the reverse model part along with their classification and probability values (see text). Each row shows images sampled from one class. Left: images obtained by recursive sampling through the net starting from the class layer till the image layer. Right: images obtained by sampling through one layer only and then recursively computing probabilities till the image layer.} 
   \label{fig:mnist-mlp-sampling}
\end{figure}

To further asses the reverse model learned by the generative approach, we sampled 10 images for each of the classes from both models. The results are shown in Fig.~\ref{fig:mnist-mlp-sampling}. For each sampled image we also applied the forward model to classify it. The classes with the highest probability along with values of that probability are shown on top of each sampled image. It is clearly seen that the generative learning approach has found a set of model parameters that yields simultaneously good classification accuracy and the ability to generate images of average digits. On the other hand, it is also visible from Fig.~\ref{fig:mnist-mlp-sampling} that the generative model part learned only the mean shapes of the digits.

\subsection{MNIST CNN} \label{subsec:mnist_cnn}
\begin{figure}[ht]
  {\centering
  \includegraphics[width=0.4\textwidth]{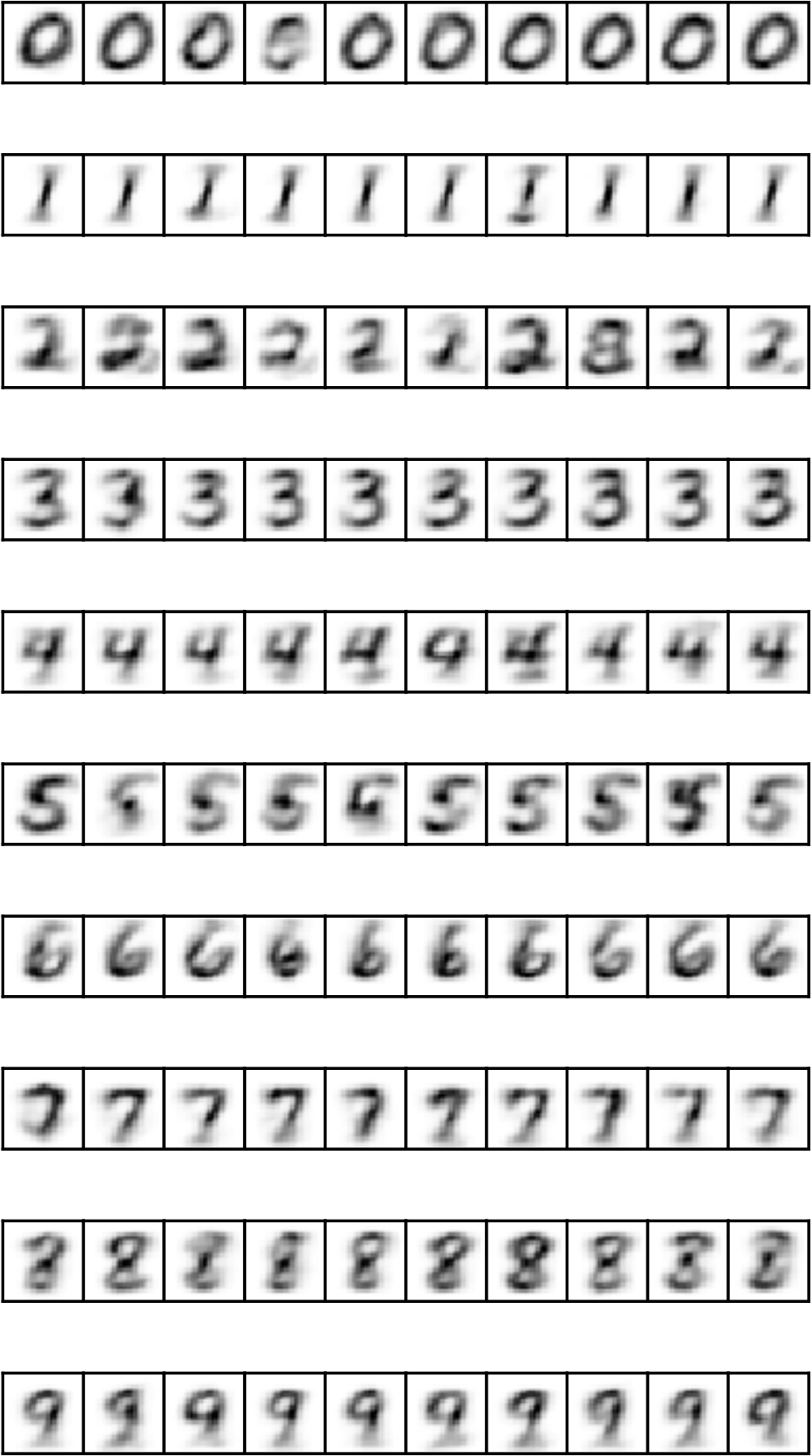} \hspace{0.05\textwidth}
  \includegraphics[width=0.4\textwidth]{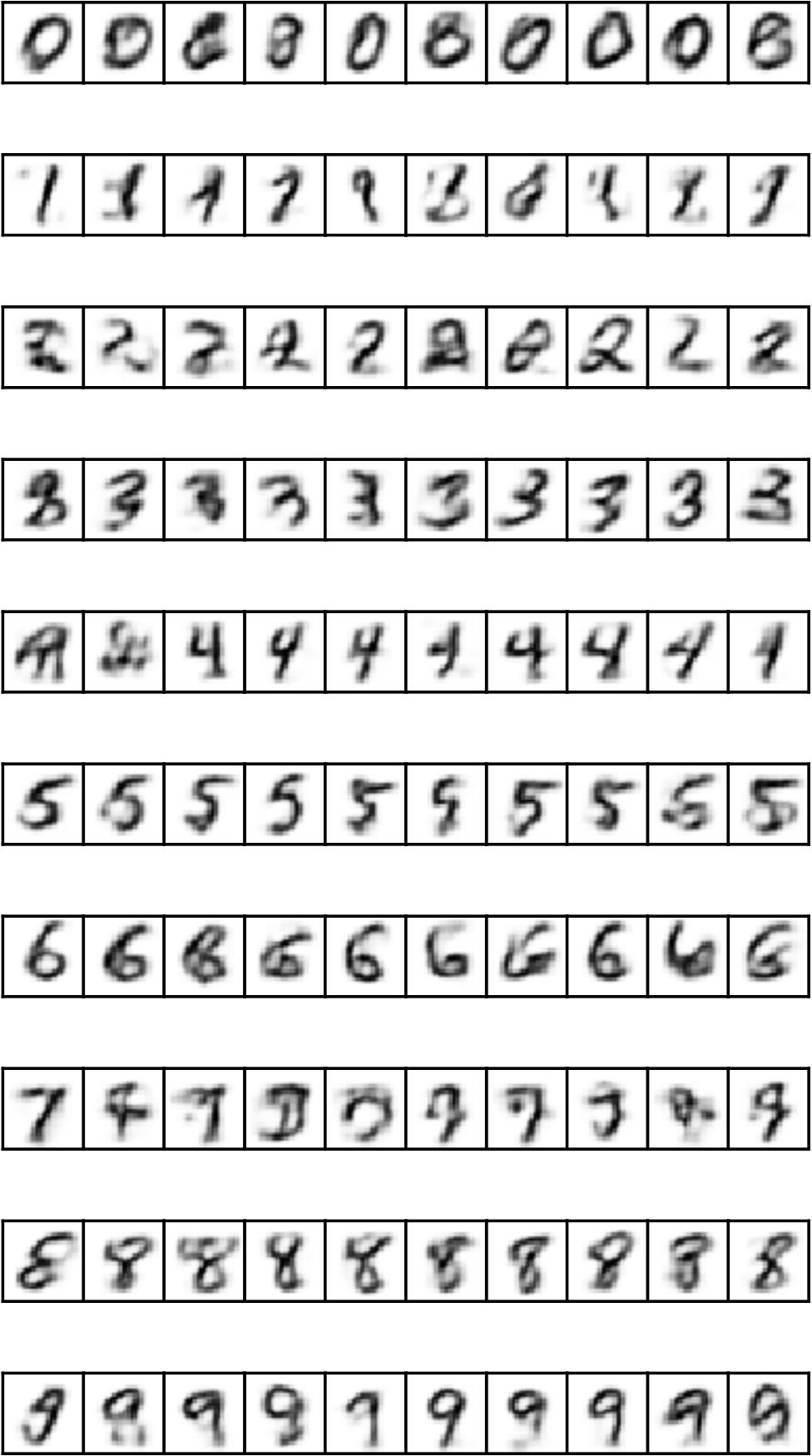}\\
  }
  \caption{Images sampled from generatively learned CNNs on MNIST data. For better visibility the images were obtained by sampling through one layer only and then recursively computing probabilities till the image layer. Left: learning method as described in Sec.~\ref{subsec:gen_learn}. Right: alternative learning method described in Sec.~\ref{subsec:mnist_cnn}.}
   \label{fig:mnist-cnn-sampling}
\end{figure}

The second experiment intends to analyse the behaviour of the proposed generative learning approach for convolutional neural networks. We learned a deep CNN ( eight layers) with architecture ((3,1,32), (3,1,32), (2,2,32), (3,1,64), (3,1,64), (2,2,64), (50), (10)) on MNIST data. The components of the triplets denote the window size, stride and kernel number for convolutional layers. The singletons denote the size of dense layers.  Sigmoid activation was used for all but the last layer. When using the objective \eqref{eq:pplik2}, the net achieves 0.989 validation accuracy for the forward part and 0.23 loss for the reverse model part. We observed neither over-fitting nor ``vanishing gradient'' effects. We conjecture that the reverse learning task serves as a strong regulariser. The left tableaux in Fig.~\ref{fig:mnist-cnn-sampling} shows images sampled from the reverse model part. Again, for better visibility, we sampled from the the distribution $q(\mvec{x}^{d-1} \mid \mvec{x}^{d})$ and then recursively 
computed the probabilities down to the image layer by applying the approximation \eqref{eq:pyapprox}. It is clearly seen that the learned generative model part is not able to capture the multi-modal image 
distribution. We conjecture that one of the possible reasons is the simplicity of the approximation \eqref{eq:pyapprox}, especially when applied for learning of the reverse model part.

To analyse the problem, we considered a somewhat different learning objective for this model part
\begin{equation} \label{eq:pplik3}
  \frac{1}{\abs{\mathcal{T}}} 
 \sum_{(\mvec{x}^0,\mvec{x}^d)\in \mathcal{T}} \bigl[
 \log p_{\theta}(\mvec{x}^d \mid \mvec{x}^0) + 
 \log q_{\theta}(\mvec{x}^0 \mid \mvec{x}^{d-1}) \bigr] 
 \rightarrow \max_{\theta}.
\end{equation}
where $\mvec{x}^{d-1}$ is sampled from $p_{\theta}(\mvec{x}^{d-1} \mid \mvec{x}^0)$ for the current $\theta$. The model learned with this objective achieved 0.990 validation accuracy and 0.14 validation loss for the reverse model part. Images sampled from this model in the same way as described above, are shown in the right tableaux of Fig.~\ref{fig:mnist-cnn-sampling}. It is clearly seen that this model captured modes in the image distribution, i.e.~different writing styles of the digits.

All experiments presented in this paper were carried out by using Keras \cite{chollet:keras2015} on top of Tensorflow \cite{tensorflow2015a}. 
\section{Conclusions}
We proposed a class of probabilistic models that leads to a new statistical interpretation of DNNs in which all neurons of the net become random variables. In contrast to Boltzmann machines, this class allows for an efficient approximation when computing the distribution of the output variables conditioned on a realisation of the input variables. The computation itself becomes identical to the recursive forward computation for a DNN.

The second objective of the paper, to design a generative learning approach for DCIMs, has been reached only partially. The proposed approach and its variants do allow to learn the forward and backward model parts simultaneously. However, the achievable expressive power of the backward part of the model is still inferior to GANs.

Notwithstanding this, we believe that the presented approach opens interesting research directions. One of them is to analyse the proposed approximation and to search for better ones, leading to new and better activation functions. Another question in that direction is a statistical interpretation of the ReLu activation as the expectation of a possibly continuous random variable.

A different direction calls for generalising DCIMs by introducing interactions between neurons in layers. This would lead to deep CRFs. We envision to learn such models, at least discriminatively, by using e.g.~recent results for efficient marginal approximations of log-supermodular CRF.

\subsubsection*{Acknowledgements}
A.~Shekhovtsov was supported by Toyota Motor Europe HS, O.~Fikar was supported by Czech Technical University in Prague under grant SGS17/185/OHK3/3T/13 and B.~Flach was supported by Czech Science Foundation under grant 16-05872S.

\bibliographystyle{apa}
\bibliography{bib/neuro-generative}

\begin{thebibliography}{}

\bibitem[\protect\astroncite{Abadi et~al.}{2015}]{tensorflow2015a}
Abadi, M. et~al. (2015).
\newblock {TensorFlow}: Large-scale machine learning on heterogeneous systems.
\newblock Software available from tensorflow.org.

\bibitem[\protect\astroncite{Chollet et~al.}{2015}]{chollet:keras2015}
Chollet, F. et~al. (2015).
\newblock Keras.
\newblock \url{https://github.com/fchollet/keras}.

\bibitem[\protect\astroncite{Goodfellow et~al.}{2014}]{Goodfellow:NIPS2014}
Goodfellow, I.~J., Pouget{-}Abadie, J., Mirza, M., Xu, B., Warde{-}Farley, D.,
  Ozair, S., Courville, A.~C., and Bengio, Y. (2014).
\newblock Generative adversarial nets.
\newblock In {\em Advances in Neural Information Processing Systems 27: Annual
  Conference on Neural Information Processing Systems 2014, December 8-13 2014,
  Montreal, Quebec, Canada}, pages 2672--2680.

\bibitem[\protect\astroncite{Hinton et~al.}{1995}]{HintonDayanFreyEtAl95}
Hinton, G.~E., Dayan, P., Frey, B.~J., and Neal, R.~M. (1995).
\newblock The wake-sleep algorithm for unsupervised neural networks.
\newblock {\em Science}, 268:1158.

\bibitem[\protect\astroncite{Hinton et~al.}{2006}]{Hinton:NC2006}
Hinton, G.~E., Osindero, S., and Teh, Y.~W. (2006).
\newblock A fast learning algorithm for deep belief networks.
\newblock {\em Neural Computation}, 18(7):1527--1554.

\bibitem[\protect\astroncite{Kingma and Welling}{2013}]{KingmaW13}
Kingma, D.~P. and Welling, M. (2013).
\newblock Auto-encoding variational bayes.
\newblock {\em CoRR}, abs/1312.6114.

\bibitem[\protect\astroncite{Kingma and Welling}{2014}]{Kingma:2014}
Kingma, D.~P. and Welling, M. (2014).
\newblock Efficient gradient-based inference through transformations between
  bayes nets and neural nets.
\newblock In {\em Proceedings of the 31st International Conference on
  International Conference on Machine Learning - Volume 32}, ICML'14, pages
  II--1782--II--1790. JMLR.org.

\bibitem[\protect\astroncite{LeCun and Cortes}{2010}]{LeCun:MNIST2010}
LeCun, Y. and Cortes, C. (2010).
\newblock {MNIST} handwritten digit database.

\bibitem[\protect\astroncite{Lee et~al.}{2009}]{Lee:ICML2009}
Lee, H., Grosse, R., Ranganath, R., and Ng, A.~Y. (2009).
\newblock Convolutional deep belief networks for scalable unsupervised learning
  of hierarchical representations.
\newblock In {\em Proceedings of the 26th Annual International Conference on
  Machine Learning}, ICML '09, pages 609--616, New York, NY, USA. ACM.

\bibitem[\protect\astroncite{Mnih and Gregor}{2014}]{Mnih-2014}
Mnih, A. and Gregor, K. (2014).
\newblock {Neural Variational Inference and Learning in Belief Networks}.
\newblock In {\em {Proceedings of the 31st International Conference on Machine
  Learning, Cycle 2}}, volume~32 of {\em {JMLR Proceedings}}, pages 1791--1799.
  {JMLR.org}.

\bibitem[\protect\astroncite{Neal}{1992}]{Neal:1992}
Neal, R.~M. (1992).
\newblock Connectionist learning of belief networks.
\newblock {\em Artif. Intell.}, 56(1):71--113.

\bibitem[\protect\astroncite{Radford et~al.}{2016}]{Radford:ICLR2016}
Radford, A., Metz, L., and Chintala, S. (2016).
\newblock Unsupervised representation learning with deep convolutional
  generative adversarial networks.
\newblock In {\em ICLR}.

\bibitem[\protect\astroncite{Rezende et~al.}{2014}]{Rezende:ICML2014}
Rezende, D.~J., Mohamed, S., and Wierstra, D. (2014).
\newblock Stochastic backpropagation and approximate inference in deep
  generative models.
\newblock In Jebara, T. and Xing, E.~P., editors, {\em Proceedings of the 31st
  International Conference on Machine Learning (ICML-14)}, pages 1278--1286.
  JMLR Workshop and Conference Proceedings.

\bibitem[\protect\astroncite{Rumelhart et~al.}{1986}]{Rumelhart:Nature1986}
Rumelhart, D.~E., Hinton, G.~E., and Williams, R.~J. (1986).
\newblock Learning representations by back-propagating errors.
\newblock {\em Nature}, 323:533--536.

\bibitem[\protect\astroncite{Salakhutdinov}{2015}]{Salakhutdinov:ARSA2015}
Salakhutdinov, R. (2015).
\newblock Learning deep generative models.
\newblock {\em Annual Review of Statistics and Its Application}, 2:361--385.

\bibitem[\protect\astroncite{Salakhutdinov and Hinton}{2009}]{salakhutdinov09a}
Salakhutdinov, R. and Hinton, G.~E. (2009).
\newblock {D}eep {B}oltzmann {M}achines.
\newblock In {\em AISTATS}, pages 448--455.

\end{thebibliography}


\end{document}